\documentclass[letterpaper]{IEEEtran}
\IEEEoverridecommandlockouts
\usepackage{cite}
\usepackage{amsmath,amssymb,amsfonts}
\usepackage{graphicx}
\usepackage{textcomp}
\usepackage{xcolor}
\usepackage{verbatim}
\usepackage{caption}
\usepackage{subcaption}
\usepackage{algorithm}
\usepackage{algpseudocode}
\usepackage{paralist}
\usepackage{tikz}
\usepackage{url}
\usepackage{hyperref}

\newcommand\copyrighttext{%
  \footnotesize This is the preprint accepted for publication in the 19th International Conference on Distributed Computing in Smart Systems and the Internet of Things (DCOSS-IoT), Pafos, Cyprus, June 19-21, 2023. This version is released under a CC-BY license according to the requirements of the Horizon Europe programme that has provided funding for this work. The final version is available at \href{https://doi.org/10.1109/DCOSS-IoT58021.2023.00043}{https://doi.org/10.1109/DCOSS-IoT58021.2023.00043}.}
\newcommand\copyrightnotice{%
\begin{tikzpicture}[remember picture,overlay]
\node[anchor=north,yshift=-10pt] at (current page.north) {\fbox{\parbox{\dimexpr\textwidth-\fboxsep-\fboxrule\relax}{\copyrighttext}}};
\end{tikzpicture}%
}

\def\BibTeX{{\rm B\kern-.05em{\sc i\kern-.025em b}\kern-.08em
    T\kern-.1667em\lower.7ex\hbox{E}\kern-.125emX}}

\begin{document}

\title{Flexible Computation Offloading at the Edge for Autonomous Drones with Uncertain Flight Times\\
}

\author{\IEEEauthorblockN{Giorgos Polychronis and Spyros Lalis} \\
\IEEEauthorblockA{\textit{Electrical and Computer Engineering Department} \\
\textit{University of Thessaly}\\
Volos, Greece \\
\{gpolychronis, lalis\}@uth.gr %or ORCID
}
}

\maketitle

\copyrightnotice

\begin{abstract}
An ever increasing number of applications can employ aerial unmanned vehicles, or so-called drones, to perform different sensing and possibly also actuation tasks from the air. In some cases, the data that is captured at a given point has to be processed before moving to the next one. Drones can exploit nearby edge servers to offload the computation instead of performing it locally. However, doing this in a naive way can be suboptimal if servers have limited computing resources and drones have limited energy resources. 
In this paper, we propose a protocol and resource reservation scheme for each drone and edge server to decide, in a dynamic and fully decentralized way, whether to offload the computation and respectively whether to accept such an offloading requests, with the objective to evenly reduce the drones' mission times. We evaluate our approach through extensive simulation experiments, showing that it can significantly reduce the mission times compared to a no-offloading scenario by up to $26.2\%$,  while outperforming an offloading schedule that has been computed offline by up to $7.4\%$ as well as a purely opportunistic approach by up to $23.9\%$.
\end{abstract}


\section{Introduction}
\label{sec:intro}

Thanks to the recent developments in embedded control systems and autopilot software, aerial unmanned vehicles, so-called drones, have become popular sensor/actuator platforms for a wide range of civilian application. Polycopters, in particular, are able to perform vertical take-off and landing without requiring a lot of space, and can steadily hover at a given position, which makes them especially attractive for smart city, security and search-and-rescue operations.

There are several application scenarios where the drone needs to scan an area by visiting specific locations to perform certain sensing tasks. Moreover, the drone may need to process this data on the spot, e.g., to detect objects of interest, before moving to the next location. While such processing can be performed using an onboard computer, for heavyweight computations, this can take a long time given the limitations of such embedded platforms. Instead, edge computing can be used to accelerate processing thereby leading to reduced mission times and greater operational autonomy. 

However, planning such missions and scheduling computation offloading accordingly, becomes non-trivial if the available edge servers need to be used concurrently by several drones that operate in the area, but have limited capacity and thus cannot execute the required computations in parallel. The problem becomes more challenging taking into account the typically limited flight time of drones, especially if they are battery-powered. In this case, failure to offload the computation to an edge server as planned, not only increases the mission time but may also necessitate changes in the planned path so that the drone can perform an intermediate stop to a depot station in order to switch batteries. Last but not least, the time needed for a drone to fly between two points may not be known with full certainty when the mission is being planned, thus the actual flight times may be different than those assumed when constructing the mission plan. 

To tackle the above problem, we propose an offloading protocol that takes as input an initial mission plan (produced offline) but allows each drone to exploit the available edge server infrastructure in a flexible and fair way so that the mission time is reduced evenly for all drones. The main contributions of this work as as follows:
\begin{inparaenum} [(i)]
\item we capture the salient aspects of the problem at hand using a formal system model;
\item we describe a protocol allowing autonomous drones to take offloading decisions; 
\item the protocol is decentralized, and does not involve a global coordinator nor does it require any interaction between drones;
\item we evaluate the protocol through extensive simulation experiments, using realistic system parameters;
\item we show that the proposed approach can significantly reduce the mission time of all drones in an even way; 
\item in fact, our approach 
outperforms mission execution based on the offline plan as well as vs a pure opportunistic offloading approach, by up to $7.4\%$ and $23.9\%$, respectively.
\end{inparaenum}

The rest of the paper is structured as follows. Section~\ref{sec:model} provides the system model. In Section~\ref{sec:approach}, we  outline our approach, which is described in detail in Section~\ref{sec:protocol}. Section~\ref{sec:eval} presents an evaluation of the proposed approach. Section~\ref{sec:related} gives an overview of related work. Finally, Section~\ref{sec:concl} concludes the paper.

\section{System Model}
\label{sec:model}

This section captures in a formal way the mission model and the key overheads related to its execution which make-up the total mission time. The model is used on the one hand to clearly formulate the optimization problem we addressed in this work, and on the other hand to allow the drone to take offloading decisions accordingly.  

\subsection{Drones and flight paths}

We assume $M$ drones, $d_m, 1 \leq m \leq M$, which are used concurrently to support different applications. To this end, each drone is assigned an independent mission, in which it must visit certain points of interest $\mathcal{V}_m$. Each such point represents a location where the drone needs to perform some sensing and process the data in-situ, e.g., to take further action if a problem is detected, before moving to the next point. 
Also, each drone $d_m$ starts its mission by taking off from a depot station $dep_m$, which is not itself a point of interest ($dep_m \notin \mathcal{V}_m$), and ends its mission by returning back and landing there. 

To visit the assigned points of interest, each drone $d_m$ follows a path $P_m$, 
encoded as a sequence of waypoints 
$P_m[i], 1 \leq i \leq \text{len}(P_m)$. 
Note that $P_m[1], P_m[\text{len}(P_m)] = dep_m$ since the drone starts and ends its mission at its depot. 

\subsection{Flight behavior}

Our work focuses on polycopter drones, which have vertical take-off/landing capability 
and can steadily hover at a given position.
Let $cruiseT_m^{i,i+1}$ be the time needed for $d_m$ to fly between $P_m[i]$ and $P_m[i+1]$ at cruising altitude. Further, let $takeoffT_m$ be the time needed for $d_m$ to perform a vertical take-off from $dep_m$ and reach the cruising altitude, and let $landT_m$ be the time needed to vertically land back at $dep_m$. Taking into account these extra overheads, the total flight time between two waypoints is:
\begin{equation}
\label{eq:flyT}
\begin{array}{rcl}
    flyT_m^{i,i+1} & = & cruiseT_m^{i,i+1} + extraT_m^{i,i+1}\\
    extraT_m^{i,i+1} & = &
    \begin{cases}
        0, & P_m[i], P_m[i+1] \neq dep_m\\
        takeoffT_m, & P_m[i] = dep_m \\
        landT_m, & P_m[i+1] = dep_m
    \end{cases}
\end{array}
\end{equation}
The case $P_m[i] = P_m[i+1] = dep_m$ is not handled in the above equation because this never occurs in a properly-formed path; it would mean that the drone takes-off from the depot only to immediately land back. 

Note that $cruiseT_m^{i,i+1}$ can be calculated as a function of the geographical distance between $P[i]$ and $P[i+1]$,  the drone's horizontal acceleration/deceleration and cruising speed. Similarly, $takeoffT_m$ and $landT_m$ can be estimated based on the vertical acceleration/deceleration of the drone and the cruising altitude for the mission at hand, or simply by measuring the respective times for take-off and landing to/from the cruising altitude. 

\subsection{Sensing and computation time}

When a drone arrives at a point of interest, it takes measurements via its sensors (e.g., pictures using an onboard camera) and then processes this data. %A given drone always performs the same type of sensing and data processing at each point of interest. But different drones may perform different types of sensing and data processing. 
Let $senseT_m$ be the time needed for $d_m$ to perform the required sensing. 
Also, let $comp_m$ be the computation to be performed, 
taking $data_m^{in}$ amount of data as input and returning $data_m^{out}$ amount of data as a result.

Each drone $d_m$ carries an onboard hardware platform $hw_m$ with sufficient resources to perform the computation at hand. 
Let $procT(comp_m,hw_m)$ be the time needed to perform the computation on the local hardware platform. We assume that the input/output data transfer time to/from the local computing hardware is negligible. Thus the time needed for the results to be ready is equal to the processing time.

As another option, the drone can offload its computation to a nearby edge server in order to accelerate processing. 
We assume that edge servers have strong computing platforms and are prepared to provide such a computation as a service, by shipping to them the respective micro-services (e.g., in the form of containers) before the drones start their missions. 
Let there be $K$ edge servers, $s_k, 1 \leq k \leq K$,  
at various locations in the area where the drones operate. Each server $s_k$ has a hardware platform $hw_k$, 
and is accessible via a dedicated local wireless network with bandwidth $bw_k$ and communication range $range_k$. When drone $d_m$ visits a point of interest $P_m[i]$ which is in range of $s_k$, it can offload $comp_m$ to it. Then, the time needed for the drone to receive the results is
\begin{equation}
\label{eq:compT}
     compT(m,k) = procT(comp_m,hw_k) + \frac{data_m^{in} + data_m^{out}}{bw_k}
\end{equation}
taking into account the processing time on the server and the time needed to transfer the input/output data over the network. 
Let $S_m[i]$ encode the offloading for $d_m$ during its mission. More specifically, $S_m[i]=k$ if the drone shall offload $comp_m$ to $s_k$ at $P_m[i]$, else $S_m[i]=0$ if the drone shall perform the computation locally. 

In the general case, the drone may not be able to \emph{immediately} offload its computation to $S_m[i]$ as soon as it completes its sensing task at  
$P_m[i]$, because the server's resources may be used by other drones at that time. However, it may still be beneficial for the drone to wait for the server to become available and then offload the computation as usual, instead of performing the computation locally. Let the extra waiting time for this be $W_m[i] \geq 0$.   
Based on all the above, the total time spent by the drone to perform the necessary sensing and computation at each point of interest 
is 
\begin{equation}
\label{eq:wpT}
visitT_m^i = senseT_m + W_m[i] + compT(m,S_m[i]), P[i] \in \mathcal{V}_m 
\end{equation}
Also, in order for the above equation to capture the case where $S[i]=0$ and the computation is performed locally, we let $compT(m,0)=procT(comp_m,hw_m)$ and  $W_m[i]=0$.

\subsection{Energy consumption and energy renewal}

We assume drones with electric motors powered via a battery\footnote{Our model can also be applied to drones featuring a conventional fuel engine and a tank with limited fuel capacity that may need to be refilled}. Let $E_m^{max}$ be the energy storage capacity of $d_m$'s battery. 
Naturally, the drone consumes energy in order to execute its mission. Let the energy that is consumed by $d_m$ to fly between $P[i]$ and $P[i+1]$ be a linear function of the corresponding flight time %$flyE_m^{i,i+1} = 
$\beta_m \times flyT_m^{i,i+1}$.  
In the same spirit, the energy that is consumed by the drone to hover above $P[i]$ is a linear function of the time spent to perform the required sensing and computation %$pE_m^i = 
$\gamma_m \times 
visitT_m^i, P[i] \in \mathcal{V}_m$.

Each time the drone flies between $P[i]$ and $P[i+1]$ or visits a point of interest $P[i] \in \mathcal{V}_m$ it reduces its energy reserves $e_m^{new} = e_m^{old} - \beta \times flyT_m^{i,i+1}$ or $e_m^{new} = e_m^{old} - \gamma \times visitT_m^{i}$, respectively. Note that the drone should never run out of energy in mid-air, i.e., $e_m^{new} > 0$ should hold at all times during the mission. Otherwise the drone would be forced to make an emergency landing, which is undesirable for many reasons including the risk for damages or even injuries.

To satisfy this constraint, drones with limited autonomy or long missions may have to perform intermediate depot stops (detours) in order to switch batteries before they can proceed with the rest of their mission. Such detours are encoded in the drone's path in an explicit way, by inserting $P[i]=dep_m$ as an intermediate waypoint between two points of interest $P[i-1],P[i+1] \in \mathcal{V}_m$. Note that Equation~\ref{eq:flyT} is sufficiently generic to capture the flight times for such detours. 

Further, let $depT_m$ be the time it takes for $d_m$ to switch batteries (restoring its energy reserves to $e_m^{new} = E_m^{max}$) once it has landed at its depot $dep_m$. For this special case, which is not captured by Equation~\ref{eq:wpT}, we set 
\begin{equation}
\label{eq:wpT_extended}
    visitT_m^i=
    \begin{cases}
        depT_m, & P_m[i] = dep_m \land 1 < i < \text{len}(P_m) \\
        0, & P_m[i] = dep_m \land (i = 1 \lor i = \text{len}(P_m))
    \end{cases}
\end{equation}
The first case captures the time spent at the depot to switch the drone's batteries.
In the second case where the drone is at its depot in the beginning or end of its path, the time spent at the depot is $0$. This is because each drone starts its mission with fresh batteries and the mission is considered to be completed the moment the drone lands at its depot for the last time (no battery switch is required). 

\subsection{Mission plan and mission time}

The mission plan for drone $d_m$ can be expressed as a triplet $(P_m, S_m, W_m)$, which consists of the flight path $P_m$ to be followed along with the respective offloading schedule $S_m$ and corresponding waiting times $W_m$. Based on all the above, the time needed to execute the plan is
\begin{equation}
\label{eq:missionT}
missionT_m = visitT^1_m + \sum_{i=1}^{\text{len}(P_m)-1}{flyT_m^{i,i+1} + visitT_m^{i+1}} 
\end{equation}
The terms $visitT_m^1$ and 
$visitT_m^{\text{len}(P_m)}$ are included for completeness but do not affect the mission time since they are $0$ (Equation~\ref{eq:wpT_extended}).

\begin{figure*}[!h]
    \centering
    \includegraphics[width=1.0\textwidth]{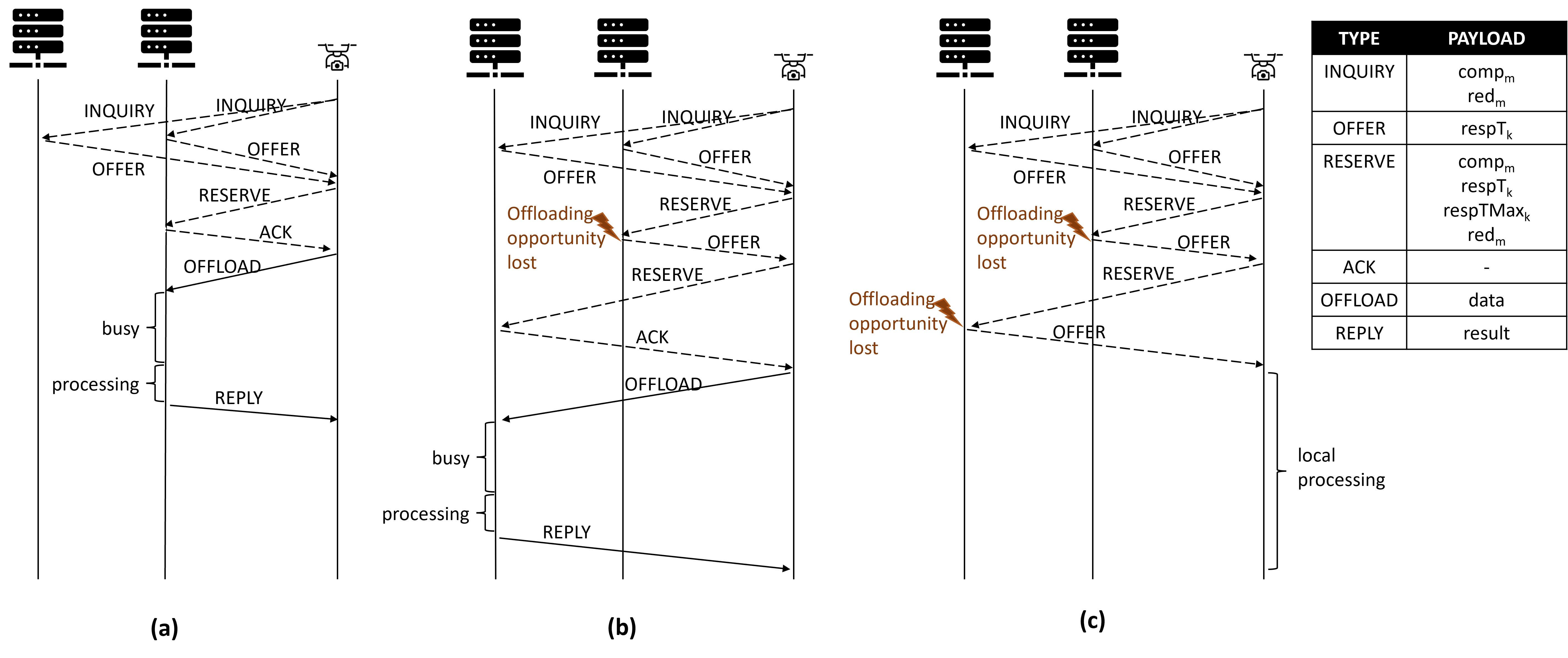}
    \caption{Indicative scenarios for offloading the drone's computation to the edge. The messages of the ad-hoc protocol are shown with dashed arrows. The solid arrows show the actual offloading interaction (if the drone decides to engage the edge server).}
    \label{fig:protocol}
\end{figure*}

\section{Objective and Approach}
\label{sec:approach}

Our objective is to exploit the available edge server infrastructure in order to reduce the mission times of all drones that execute their mission concurrently, in a fair manner. %More specifically, the problem we tackle in this work is the following: 
%Given drones $d_m, 1 \leq m \leq M$ with assigned points of interest $\mathcal{V}_m$ where they have to perform some sensing and data processing, and stationary edge servers $s_k, 1 \leq k \leq K$ located in the mission area, which can be used for offloading the drones' computations, produce mission paths $P_m$ and offloading plans $S_m, W_m$ that are feasibly energy-wise and minimize $mT_m$ (as per Equation~\ref{eq:missionT}) \emph{in a fair way for all drones}. 
We capture fairness using as a reference a default mission plan $(P_m^{def}, S_m^{def}, W_m^{def})$, where $P_m^{def}$ is produced by a TSP algorithm without considering offloading ($S^{def}_m[i], W^{def}_m[i]=0, 1 \leq i \leq \text{len}(P_m^{def})$) or energy constraints, and where depot detours are subsequently added in the path as needed to ensure that the drone will not run out of energy. Then, the relative reduction of the default mission time that is achieved by an alternative plan $(P_m, S_m, W_m)$ is 
\begin{equation}
\label{eq:red}
red_m = \frac{missionT_m^{def} - missionT_m}{missionT_m^{def}} 
\end{equation}
where $missionT_m^{def}$ and $missionT_m$ are the respective mission times, as per Equation~\ref{eq:missionT}. 
Our objective is to
\begin{equation}
\label{eq:opt}
\text{maximize } min_{m=1}^{M}(red_m) 
\end{equation}
In other words, %using the available server infrastructure for offloading, 
we wish for each drone to execute its mission so as to maximize the worst relative reduction of the default mission time among all drones. 

%We assume that the drone's maximum battery capacity $E_m^{max}$ is sufficient so that $d_m$ can move from its depot $dep_m$ (starting with full batteries) to any point of interest $P_m[i]$ and back to $dep_m$ even if the computation at that point is performed locally on the drone. This is to ensure that the problem always has a feasible solution irrespective of the points of interest assigned to each drone and the degree of contention among drones for the shared server resources.

In previous work~\cite{polychronis22}, we have proposed an offline heuristic that takes as input the default mission plan $(P^{def}_m, S^{def}_m, W^{def}_m)$ for each drone $d_m$ and generates an optimized plan $(P_m, S_m, W_m)$ for the above objective. %At runtime, the drones simply follow this plan. 
However, his approach assumes that the flight times $flyT_m^{i,i+1}$ as per  Equation~\ref{eq:flyT} are known a priori with high accuracy. This can be unrealistic in case of variable weather conditions or random wind gusts due to the morphology of the mission area.     

In this work, the flight times between two waypoints are not known with certainty during the planning phase. Instead, we assume that the actual flight times vary according to a uniform random distribution with upper bound $flyMaxT^{i,i+i}_m$ and lower bound $flyMaxT^{i,i+i}_m$. Note that this randomness directly affects the corresponding energy consumption. %, let $flyMaxE^{i,i+1}_m$ and respectively $flyMinT_m^{i,i+1}$. 
To deal with this uncertainty, we use as a starting point the mission plan produced by the offline algorithm for the worst-possible flight times and energy consumption, and propose an ad-hoc protocol for deciding, at runtime, whether and to which edge server each drone shall offload its computation at each point of interest $P[i] \in \mathcal{V}_m$. As a result, the drone may pick a different offloading option than the one suggested by the offline plan. In addition, any time/energy savings achieved vs the offline plan are exploited to postpone or even eliminate depot detours so as to further reduce the mission times. 

It is important to stress that the proposed approach is fully decentralized and does not require any coordinator nor any peer-to-peer interaction between drones. As a consequence, it can scale to a very large number of drones and edge servers with practically negligible runtime overhead for the drone.       
\section{Offloading Protocol}
\label{sec:protocol}

Each drone $d_m$ executes its mission based on the plan $(P_m S_m, W_m)$ that is generated by the offline algorithm. However, when the drone reaches a point of interest where the plan is to offload its computation to the edge ($S_m[i] \neq 0$), it runs a protocol with all edge servers in range to decide whether to offload its computation to one of them. The protocol is illustrated in Figure~\ref{fig:protocol}, for indicative offloading scenarios.  

As a first step, the drone sends an INQUIRY message to all servers in range, which includes the type of the computation at hand $comp_m$ and the expected relative reduction $red_m$ vs the default mission time. In turn, each server $s_k$ replies with an OFFER message, which indicates the estimated response time $respT_k$. This is calculated based on the processing time on the server's hardware platform $procT(comp_m,hw_k)$, the current load of the server as well as any pending offloading requests in a local priority queue. More specifically, if $d_m$ has a worse expected reduction $red_m$ compared to other pending requests, the server will give priority to it (in line with the optimization objective). Note, however, that the offer made by the server is non-binding as no actual resource allocation takes place and no entry is added in the priority queue at that point. 

When the drone receives the offers from the servers, it picks the server $s_k$ that replied with the best offer (shortest $respT_k$), and calculates the expected end-to-end computation time $compT(m,k)$ as per Equation~\ref{eq:compT} by replacing $procT(comp_m, hw_k)$ with $respT_k$.  If this is beneficial vs the local execution of the computation $procT(comp_m, hw_m)$, it sends to $s_k$ a RESERVE message to make an actual reservation for the offered response time $respT_k$. In this message, the drone also includes an upper bound for the response it is willing to accept (in case the server needs to prioritize other drones, whose requests may arrive in the near future). This is calculated as $respTMax_k = max(respT_k, visitT_m^i - senseT_m)$, where $visit_m^i - senseT_m$ is the expected offloading time in the offline plan at $P_m[i]$. If the server can indeed meet $respT_k$, it makes the necessary resource allocation, adds the request of $d_m$ in the priority queue, and replies with an ACK message, which represents a hard commitment to perform the computation by the indicated maximum response time. In this case, shown in Figure~\ref{fig:protocol}(a), the drone proceeds to offload the computation to $s_k$ and waits for the response in oder to continue its mission.

However, at the time $s_k$ receives the RESERVE message from $d_m$, it may have already committed its resources to 
another drone that run the same protocol concurrently to $d_m$. In this case, the server replies with another OFFER message and an updated $respT_k$. Based on the new offer, the drone repeats the selection process. As a result, it may decide for another edge server to which it will send a RESERVE message. If that server replies with an ACK, the drone will offload its computation to it, as shown in Figure~\ref{fig:protocol}(b). 

The drone repeats the inquiry-reservation procedure until a beneficial offloading option is successfully reserved, or the best offer received is not beneficial vs local computation. In the latter case, shown in Figure~\ref{fig:protocol}(c), the drone decides to perform the computation locally.

Algorithm~\ref{alg:main} shows the high level mission execution logic of the drone. More precisely, each drone follows the path that was generated by the offline algorithm. However, before starting the trip from the current location $P_m[i]$ to the next point of interest $P[i+1] \in \mathcal{V}_m$, a check is made to ensure that it is safe to make the flight and perform the required sensing and processing at the point of interest, based on the worst possible scenario where the drone experiences the maximum flight time and ends-up performing the computation locally. If the remaining energy of the drone is not sufficient, a depot detour is added at that point (via the \emph{insertDetour()} function). Note that this check is needed even though the offline algorithm already produces the plan based on the worst-case flight times. The reason is that, since the drone may not fully respect the offloading actions of the offline plan, it is possible for the contention on the edge servers to lead to larger delays than in the offline plan, thereby rendering the path unsafe. 
The check is not needed if $P_m[i+1]=dep_m$ as in this case the drone is heading directly to the depot anyway.

If the hop is safe, the drone moves to $P[i+1]$ and (if this is a point of interest rather than the depot) performs the required sensing task. Then, if $S_m[i] \neq 0$, it runs the protocol described above. 
Function \emph{protocol()}  abstracts the communication with the nearby edge servers, returning the identifier of the server where the computation should be offloaded. 
In case no beneficial offloading option is found, the function returns $0$, indicating that the computation should be performed locally. Note that the computation is always performed locally if $S_m[i]=0$ according to the offline plan.

\begin{algorithm}[!ht]
\caption{Drone runtime logic}
\label{alg:main}
\begin{algorithmic}

\Function{DroneSide}{$P_m, S_m, W_m, missionT_m^{def}$}
    \State $missionT_m \gets \text{calcMissionT}(P_m, S_m, W_m, 1)$
    \State $red_m \gets \text{calcReduction}(missionT_m^{def}, missionT_m)$
    \State $t \gets 0$
    \State $i \gets 1$ 
    \State $e_{rem} \gets E_m^{max}$
    \While {$i < \text{len}(P_m)$}
       \If{$P[i+i] \in \mathcal{V}_m$} \Comment{going to a point of interest}
            \State $e_{fly} \gets \beta \times flyMaxT_m^{i,i+1}$
            \State $e_{visit} \gets \gamma \times (senseT_m + compT(m,0))$
            \If{$e_{rem} \leq e_{fly} + e_{visit}$}
                \State $\text{insertDetour}(P_m,i,dep_m)$
            \EndIf
        \EndIf
        \State $t_{beg} \gets \text{getTime}()$
        \State $\text{GOTO}(P_m[i+1])$
        \State $t_{end} \gets \text{getTime}()$
        \State $flyT_m^{i,i+1} \gets t_{end} - t_{beg}$
        \State $e_{rem} \gets e_{rem} - \beta \times flyT_m^{i,i+1}$
        \State $t \gets t + flyT_m^{i,i+1}$
        \State $i \gets i+1$
        \If{$P_m[i] \in \mathcal{V}_m$} \Comment{at a point of interest}
            \State $t_{beg} \gets \text{getTime}()$
            \State $data \gets SENSE()$
            \State $k \gets 0$
            \If{$S_m[i] \neq 0$}
                \State $srv \gets \text{getServersInRange}(P_m[i])$
                 \State $k \gets \text{protocol}(srv, comp_m, red_m)$
            \EndIf
            \If{ $k \neq 0$ }
                \State $reply \gets OFFLOAD(s_k, comp_m, data)$
            \Else
                \State $reply \gets PROCESS(data)$
            \EndIf
            \State $t_{end} \gets \text{getTime}()$
            \State $visitT_m^{i} \gets t_{end} - t_{beg}$
            \State $e_{rem} \gets e_{rem} - \gamma \times visitT_m^{i}$
        \Else
            \State $t_{beg} \gets \text{getTime}()$
            \State $SWITHBATTERY()$
            \State $t_{end} \gets \text{getTime}()$
            \State $visitT_m^{i} \gets t_{end} - t_{beg}$
            \State $e_{rem} \gets E_m^{max}$
        \EndIf
        \State $t \gets t + visitT_m^{i}$
        \State $\text{optimizePath}(P_m, e_{rem})$
        \State $missionT_m \gets t + \text{calcMissionT}(P_m, S_m, W_m, i)$  
        \State $red_m \gets \text{calcReduction}(missionT_m^{def}, missionT_m)$ 
    \EndWhile
\EndFunction

\end{algorithmic}
\end{algorithm}

This procedure continues until the drone visits the last point of interest and returns back to the depot. Notably, since the actual flight times are most likely to be shorter than the worst-case assumed in by the offine planning algorithm, the drone's  mission time and energy consumption is likely to decrease vs the offline plan. To exploit this gain, the drone checks whether it can safely optimize its path by postponing or even eliminating a depot detour (function \emph{optimizePath()}). This is done after each hop based on the remaining energy at that point. Similarly, the expected mission time and the reduction vs the default mission time is also updated after each hop. To this end, the drone tracks the actual time $t$ that has elapsed up to the current point $P[i]$, and adds to it the expected mission time from that point onward as per the offline plan, computed via function \emph{calcMisstionT()}, in the spirit of Equation~\ref{eq:missionT}, but starting from the indicated mission point rather than the beginning of the mission. The reduction vs the default mission time is then calculated via function \emph{calcReduction()} as per Equation~\ref{eq:red}. This is done irrespective of whether the drone's path has changed, as the mission time can be reduced merely due to the shorter flight times.    

\section{Evaluation}
\label{sec:eval}

We evaluate the proposed approach through simulation experiments, using realistic performance parameters from real platforms. We start by describing the experimental setup, how we select the values for the key simulation parameters. Then, we present and  discuss the results of our experiments. 

\subsection{Topology and missions}

\begin{figure}
    \centering
    \includegraphics[width=0.75\columnwidth]{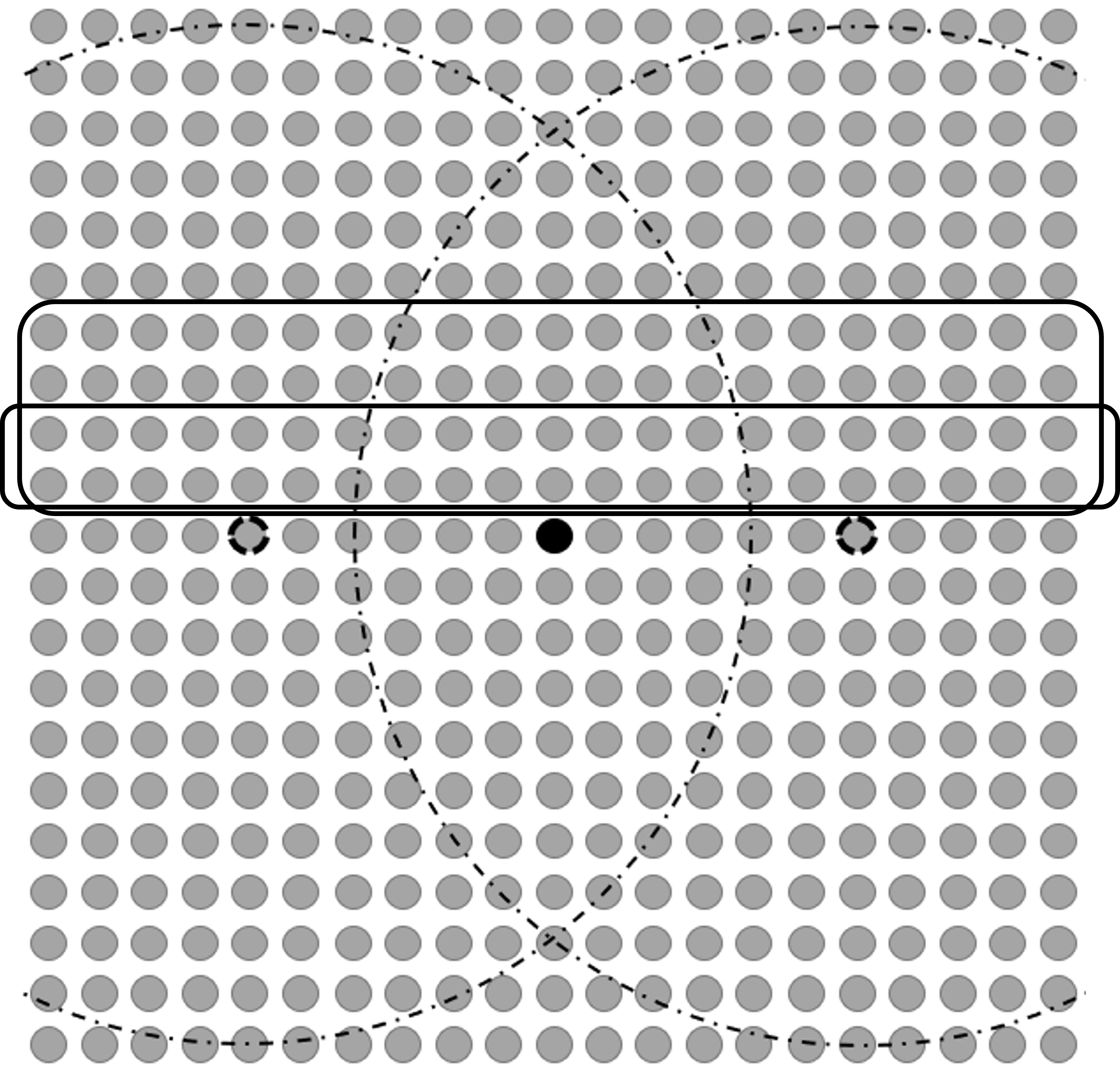}
    \caption{Mission topology.}
    \label{fig:topo}
\end{figure}

We arrange points of interest in a $21x21$ grid, shown in Figure~\ref{fig:topo}, where each node represents a possible point of interest. The neighbour nodes are set $20$ meters apart. We assume two servers located in the mission area, denoted by the gray nodes with the black borders. The dashed circles mark the range of the respective wireless networks. 
Note that some points are in range of both servers. We set the range to $200$ meters, which is feasible for line-of-sight communication using WiFi access points with proper antennas.

We run experiments for  
two mission mission types, denoted in the figure via the rounded rectangles. In the small mission 
(orange) each drone must visit $84$ points of interest, whereas in the large mission (green) each drone must visit $126$ points of interest. For symmetry, we assume a common depot station located at the center of the grid (black node), which can be used concurrently by all drones. 

\subsection{Drones}

We run experiments for $20$ drones with the same flight properties, energy capacity and hardware platform. Also, we assume that every drone performs the same sensing task and needs to perform the same type of computation on the  data. 

The cruising speed is set to $4m/s$, usual for small polycopter drones, 
while horizontal acceleration and deceleration is set to $0.8m/s^2$ and $1.6m/s^2$ respectively. Based on these parameters, the flight time between two neighbor points in the grid is about $9$ seconds. Also, we set vertical take-off and landing time, $takeoffT$ and $landT$, to $5$ and respectively $20$ seconds, assuming a cruising altitude of $10$ meters for all missions. We have confirmed this flight behavior using our own quadcopter drone in the field as well using a software-in-the-loop configuration of the drone's autopilot.

The offline algorithm produces mission plans based on the above flight behavior 
with the respective flying times assumed to be the worst case values $flyMaxT_m^{i,i+1}$. In the simulated mission execution, the flight times follow a uniform random distribution with upper bound  $flyMaxT_m^{i,i+1}$ 
and the lower bound set to $flyMinT_m^{i,i+1} = (1 - u) \times flyMaxT_m^{i,i+1}$ where $u$ is the uncertainty level. We run experiments for $u = 20\%$ and $u = 30\%$. To compare different approaches in a fair way, we store the randomly generated random flight times between every two points in a file, and then use the stored values in the simulation runs for different approaches. 

Drones are assumed to have an operational autonomy of $15$ minutes, whether cruising or hovering.  
The battery switching time at the depot $depT_m$ is set at $3$ minutes and is assumed to be correctly estimated. Thus, both in the offline plan and the simulated mission execution $visitT_m^i$ has the same value for $P[i] = dep_m$ (the invocation of the \emph{SWITHBATTERY()} function in Algorithm~\ref{alg:main} takes $depT_m$).   

As a typical sensing task, we assume that the drones use an onboard camera to take a picture. We set the image capture time $senseT$ to $1$ second, for pictures of $1$MB that are used as input $data_m^{in}$ to detect objects of interest. For the object detection computation, we use the YOLO~\cite{yolov3} software. When running this computation on a Rasberry Pi model 4B, which we consider as a typical embedded companion computer for small drones, the processing takes about $10$ seconds. This is the time used for $compT(m,0)$ both for the generation of the offline plan and the simulated execution of the mission (the invocation time of function \emph{PROCESS()} in Algorithm~\ref{alg:main}).

\subsection{Edge servers}

For the edge servers, we use a laptop with an i7-8550U CPU running a VM with $4GB$ RAM. The above object detection software is packaged as a docker container~\cite{docker}, which runs within the VM and is assumed to be preinstalled on the servers before the mission starts. We have measured the processing time to be about $1.8$ seconds. We operate the server using $1$ such micro-service, because we observed a significant increase in the processing time when two or more such services run concurrently in the VM. As a consequence, the edge server processes offloading requests in a sequential manner, one at a time. Pending requests are kept in a queue. 

The bandwidth $bw$ with the edge servers is set to $50$Mbps, which is very realistic for WiFi. Thus, the time that is required to send the offloading request (with the image) to the server 
is about $150$ milliseconds. Given that the rest of the offloading protocol messages are very small, we assume that the respective transmission time is dominated by the network latency, set to $10$ milliseconds for a WiFi link. In total, the time $compT(m,k), k \neq 0$ it takes for the drone to receive the results when the computation is offloaded to the edge is about $2$ seconds (without any additional delays due to the protocol negotiation phase or possible contention on the server). We have verified that this is very close to the actual offloading of the computation between the RPi and the laptop over WiFi. This value is used for the generation of the offline plan as well as in the simulated mission execution (this is the invocation time of function \emph{OFFLOAD()} in Algorithm~\ref{alg:main} in case the server is idle at the time when it receives the offloading request). 

\subsection{Benchmarks}

As a benchmark for the results that are achieved by our approach, we use different alternatives, briefly discussed below.

\textit{Follow the plan.} The drone follows the plan that was produced offline. More precisely, when $d_m$ arrives at $P_m[i] \in \mathcal{V}_m$ and completes the sensing task at that location, it offloads the computation to $S_m[i]$ according to plan (waiting as needed before offloading). The edge servers process requests based on a strict FCFS policy without taking into account the drone's expected reduction. Note that, in this case, there is no negotiation phase and the drone always offloads to the edge server that is specified in the plan. If $S_m[i]=0$, the drone performs the computation locally. 

\textit{Opportunistic offloading.} The drone follows the path of the offline plan. However, when $d_m$ arrives at $P_m[i] \in \mathcal{V}_m$ and completes the sensing task, it runs the protocol and takes its offloading decision irrespective of the planned $S_m[i]$ and $W_m[i]$. Notably, the drone may decide to offload even though the plan is to perform the computation locally ($S_m[i]=0$). As above, the edge servers process requests based on a strict FCFS policy without taking into account the drone's expected reduction. As in the proposed approach, the drone exploits any gains to postpone or eliminate depot detours. 

\textit{Oracle.} The mission plan is produced by the offline algorithm, based on the actual time that will be needed in the simulated mission execution to fly between two points $P_m[i]$ and $P_m[i+1]$ in the mission plan (rather than the respective worst-case values $flyMaxT_m^{i,i+1}$). The drone simply follows the path and offloading schedule of the offline plan. This basically serves as an upper bound for the result that could be achieved by the other approaches.

\begin{figure*}[h!]
\begin{center}
\subfloat[Small mission with $20$ drones.]{%
  \includegraphics[width=0.33\textwidth]{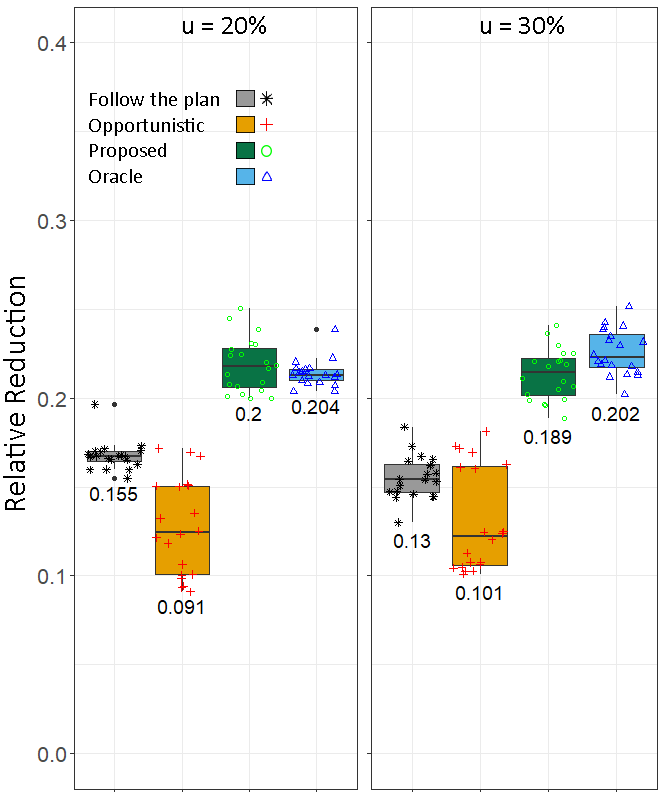}
  \label{fig:small}
} 
\hspace*{-0.5em}
\subfloat[Large mission with $20$ drones.]{%
  \includegraphics[width=0.33\textwidth]{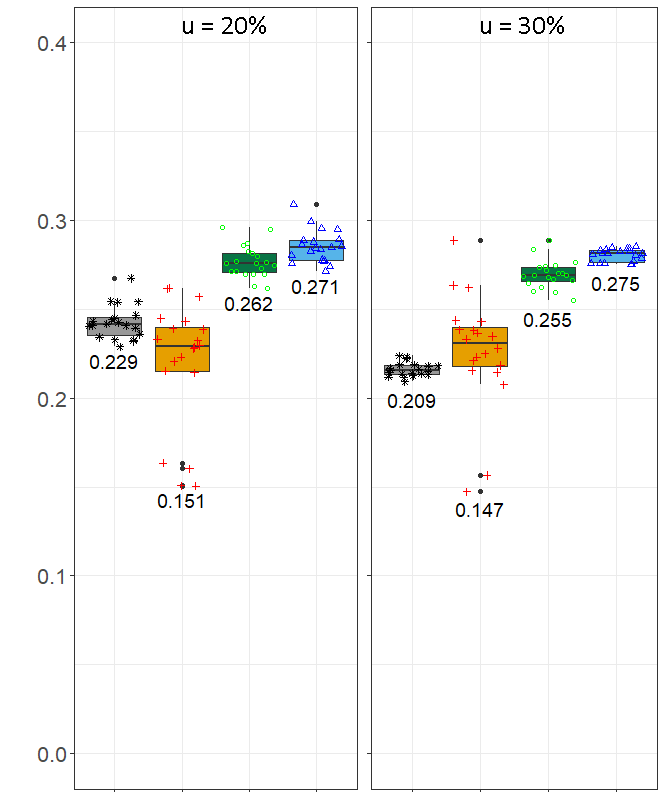}%
  \label{fig:large}
}
\hspace*{-0.5em}
\subfloat[Small and large mission, $10$ drones each.]{%
  \includegraphics[width=0.33\textwidth]{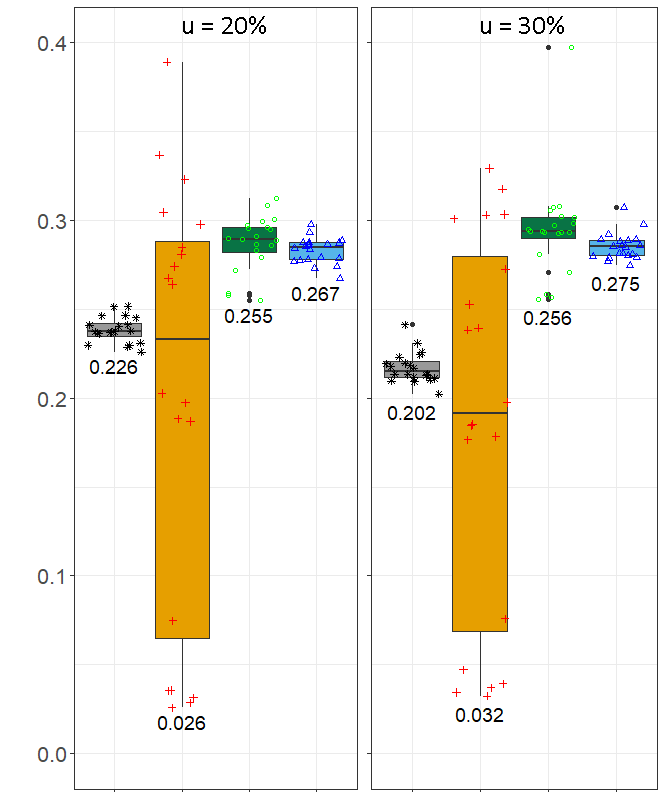}%
  \label{fig:mix}
}
\end{center}
\caption{Relative mission reduction achieved by the proposed approach vs the three benchmarks.}
\label{fig:exp}
\end{figure*}

\subsection{Results and discussion}

Figure~\ref{fig:exp} illustrates the results obtained for a particular set of randomly generated flight times. All other runs with different random values are similar. 
Figures~\ref{fig:small} and \ref{fig:large} show the results for the case where all drones follow the same mission, small and large respectively. Figure~\ref{fig:mix} shows the results for the scenario where where half of the drones follow the small mission and the other half follow the large one. The left part of each figure is for $u = 20\%$, the right part is for $u = 30\%$. The results achieved by each approach (proposed vs the three benchmarks) are illustrated as in the form of a boxplot, where  
the solid area of the box denotes the borderline to the upper and lower quartiles, the horizontal line within the area marks the median, the vertical line denotes the lower and upper quartiles, and the outliers are shown as separate dots. The markers show the relative mission time reduction achieved by each individual drone. The lowest marker (the drone with the worst relative reduction) is the actual reference for the target optimization objective, as per Equation~\ref{eq:opt}. The corresponding value is printed at the side of the marker in order for this metric to be more clearly visible. 

The proposed approach outperforms follow-the-plan in all cases. 
More precisely, it improves the worst relative reduction vs the follow-the-plan approach about $3.6\%$ and $5.4\%$ averaged over all experiments with uncertainty $20\%$ and $30\%$, respectively. In fact, this is merely $0.8\%$ and $1.2\%$ lower than the results of the oracle. 
Note that the difference  
becomes greater with increased uncertainty. This is  
expected since the latter  
does not take advantage of 
the  
shorter flight times that may occur 
during the mission. 
On the other hand, the proposed approach 
exploits such gains to optimize the path of the drone.

What is interesting in this respect, is that the proposed approach performs the same number of depot detours as the follow-the-plan approach (and the oracle) in all the cases. However,  
these depot detours are shifted to a more appropriate point in time during the mission,  
thereby reducing the total flight time vs the follow-the-plan approach,  
by $1.2\%$ and $1.3\%$  
for uncertainty $20\%$ and $30\%$, respectively.

It is important to note that the simple opportunistic approach gives constantly (far) worse results, with respect to the optimization objective, even compared to the follow-the-plan approach, especially in the mixed mission experiments (Figure~\ref{fig:mix}). This is because  
the drones try to offload all their computations to the servers,  
leading to high contention on the edge servers.
Eve though an offloading decision is taken only when this is beneficial, since the servers are jammed, the time savings are  
small for each such offloading.  
As a consequence, the aggregated time  
savings from all the computations  
are  
smaller than  
those of the offline schedule. This, in turn, has an effect in the feasibility of the  
plan. Given that the aggregated energy consumption  
at the points of interest becomes larger than estimated by the offline algorithm, at some point the path becomes unsafe, leading to the insertion of an additional depot detour; it is these detours that produce the bottom outliers 
in the figures. In contract,  
the proposed approach  
respects the offline schedule 
and does not offload unless this is part of the offline plan. Moreover, 
the servers prioritize offloading requests based on the expected reduction of the drones' mission times, rather than adopting a blind FCFS policy. 
Furthermore, any optimization of the flight path take into account the energy constraint,  
so no additional depot detour is introduced. 

\section{Related Work}
\label{sec:related}

In~\cite{kim2018dynamic}, the authors focus on the tracking of moving objects  
using a drone-based system. The goal is to dynamically decide whether to offload the computation to a server or perform it locally, so 
as to minimize the response time.
However, this work only focuses on a single drone, without considering any contention for the same edge resources due to other drones that may operate concurrently in the area.  

The authors of~\cite{chemodanov2019policy}, 
propose a scheme for deciding whether a computation will be offloaded to the edge or the cloud. Their scheme supports the need for realtime, and is modable for better performance or lower cost. In~\cite{stavrinides2021orchestrating2},  
Focus on a fog environment, where multiple workflows arrive real-time and must be offloaded to the fog. The tasks are placed in a global queue and are prioritized regarding their deadline (earliest deadline first). Then the resources in the fog or cloud are chosen based on their estimated finish time (choose the resource with the earliest finish time). In~\cite{stavrinides2021orchestrating}, workloads also arrive dynamically in the system. Then three strategies, for allocating distributed resources (processors), are evaluated. The strategies are a round robin, each time picking the next processor, a policy picking the processor with the shortest queue and two random choices policies, picking two random resources and then picking the one with the shortest queue. The authors of~\cite{chen2020intelligent} focus on minimizing the average service latency, of multiple drones requesting service from edge servers. They propose an algorithm for making offloading decisions and resource allocation.

In \cite{pham2017cost} the authors propose an algorithm for scheduling task graphs in a fog-cloud environment. The usage of the cloud comes with a financial cost. Their goal is to schedule the tasks (within the task graphs) in a way to achieve good tradeoff between performance (execution time) and cost. In their solution, the tasks of a task graph are prioritized based on their position in the task graph, then the best processing node is selected, and in the final phase tasks are reassigned to processing nodes to limit deadline violations. The authors of~\cite{tang2019offloading}, focus on a problem where moving vehicles use edge servers to offload their tasks. They propose a decentralized method to find candidate servers in an area, and then propose a deadline and (financial) cost aware approach, to choose a server. In~\cite{dong2019joint}, a game theory algorithm is proposed, where multiple users/players, that require offloading, make choices about offloading to the edge or in a central cloud. The choices of the users are coordinated in the central cloud. The objective is to minimize the cost of using the edge and the cloud while respecting the delay constraints of the services offloaded.

The authors of~\cite{peng2021dosra}, propose an approach for scheduling dynamically arriving tasks, with different priorities, to edge servers. The goal is to minimize the average priority-weighted response time. Their algorithm consists of a decentralized approach for finding a proper server  
and a strategy for relocating tasks and reallocating resources. The relocation of tasks is done so that tasks with different  priorities are executed to a appropriate edge nodes. 

A key difference with the above works, is the flight paths. On the one hand, in our work we try to optimize the total mission time, which is also affected by the flight times. On the other hand, in our approach we also adapt the path. However, similar to \cite{tang2019offloading} and \cite{peng2021dosra} we also propose an decentralized approach for our schedule adaptation. Additionally, the oportunistic offloading approach used for the comparison with our proposed algorithm, is very similar to the shortest queue approach also tested in~\cite{stavrinides2021orchestrating}. Another similarity, is that in our proposed approach also tackle the incoming inquiries with priorities, like in \cite{peng2021dosra}. However in our environment, the servers do not communicate with each other, so a task relocation strategy could not be adopted. So we tackle the problem with the usage of priority queues.

The authors of~\cite{polychronis2021safe} deal with the problem where multiple drones must visit some predefined points. However there is uncertainty regarding the flight times. 
They propose a dynamic plan adaptation that takes into account the real travel times that occurred during the mission execution and the drone's energy constraints. Our work also adapts the path dynamically, but additionally considers computation offloading on edge servers and takes corresponding resource allocation decisions.

The difference with our prior work in~\cite{polychronis22}, is the dynamic adaptation of the mission plan when there is uncertainty regarding the flight times. As shown in the evaluation, 
the offline schedule may become increasingly suboptimal in this case, 
so there is opportunity to achieve better results. The current work comes to fill this gap.  
\section{Conclusion}
\label{sec:concl}

We have presented a protocol for the flexible and dynamic offloading of computations from aerial drones to edge servers under uncertain flight times, taking into account the limited processing resources at the edge as well as the energy-constraints of drones and the need to switch batteries in order to support long missions. Our evaluation shows that the proposed approach can achieve significant improvements vs an offline mission plan as well as vs opportunistic offloading, and in fact performs close to an oracle offline algorithm that generates the mission plans using the actual flight times that are experienced at runtime.

\section*{Acknowledgments}

This work has been funded in part by the Horizon Europe research and innovation programme of the European Union, under grant agreement no~101092912, project MLSysOps.

\bibliographystyle{IEEEtran}
\bibliography{bib}

% Generated by IEEEtran.bst, version: 1.14 (2015/08/26)
\begin{thebibliography}{10}
\providecommand{\url}[1]{#1}
\csname url@samestyle\endcsname
\providecommand{\newblock}{\relax}
\providecommand{\bibinfo}[2]{#2}
\providecommand{\BIBentrySTDinterwordspacing}{\spaceskip=0pt\relax}
\providecommand{\BIBentryALTinterwordstretchfactor}{4}
\providecommand{\BIBentryALTinterwordspacing}{\spaceskip=\fontdimen2\font plus
\BIBentryALTinterwordstretchfactor\fontdimen3\font minus
  \fontdimen4\font\relax}
\providecommand{\BIBforeignlanguage}[2]{{%
\expandafter\ifx\csname l@#1\endcsname\relax
\typeout{** WARNING: IEEEtran.bst: No hyphenation pattern has been}%
\typeout{** loaded for the language `#1'. Using the pattern for}%
\typeout{** the default language instead.}%
\else
\language=\csname l@#1\endcsname
\fi
#2}}
\providecommand{\BIBdecl}{\relax}
\BIBdecl

\bibitem{polychronis22}
G.~Polychronis and S.~Lalis, ``Joint edge resource allocation and path planning
  for drones with energy constraints,'' in \emph{International Conference on
  Mobile and Ubiquitous Systems: Computing, Networking and Services
  (Mobiquitous)}.\hskip 1em plus 0.5em minus 0.4em\relax (to appear), 2022.

\bibitem{yolov3}
J.~Redmon and A.~Farhadi, ``Yolov3: An incremental improvement,'' \emph{arXiv
  preprint arXiv:1804.02767}, 2018.

\bibitem{docker}
D.~Merkel \emph{et~al.}, ``Docker: lightweight linux containers for consistent
  development and deployment,'' \emph{Linux journal}, vol. 2014, no. 239, 2014.

\bibitem{kim2018dynamic}
B.~Kim, H.~Min, J.~Heo, and J.~Jung, ``Dynamic computation offloading scheme
  for drone-based surveillance systems,'' \emph{Sensors}, vol.~18, no.~9, p.
  2982, 2018.

\bibitem{chemodanov2019policy}
D.~Chemodanov, C.~Qu, O.~Opeoluwa, S.~Wang, and P.~Calyam, ``Policy-based
  function-centric computation offloading for real-time drone video
  analytics,'' in \emph{2019 IEEE International Symposium on Local and
  Metropolitan Area Networks (LANMAN)}.\hskip 1em plus 0.5em minus 0.4em\relax
  IEEE, 2019, pp. 1--6.

\bibitem{stavrinides2021orchestrating2}
G.~L. Stavrinides and H.~D. Karatza, ``Orchestrating real-time iot workflows in
  a fog computing environment utilizing partial computations with end-to-end
  error propagation,'' \emph{Cluster Computing}, vol.~24, no.~4, pp.
  3629--3650, 2021.

\bibitem{stavrinides2021orchestrating}
------, ``Orchestrating bag-of-tasks applications with dynamically spawned
  tasks in a distributed environment,'' in \emph{2021 International Symposium
  on Performance Evaluation of Computer and Telecommunication Systems
  (SPECTS)}.\hskip 1em plus 0.5em minus 0.4em\relax IEEE, 2021, pp. 1--8.

\bibitem{chen2020intelligent}
J.~Chen, S.~Chen, S.~Luo, Q.~Wang, B.~Cao, and X.~Li, ``An intelligent task
  offloading algorithm (itoa) for uav edge computing network,'' \emph{Digital
  Communications and Networks}, vol.~6, no.~4, pp. 433--443, 2020.

\bibitem{pham2017cost}
X.-Q. Pham, N.~D. Man, N.~D.~T. Tri, N.~Q. Thai, and E.-N. Huh, ``A cost-and
  performance-effective approach for task scheduling based on collaboration
  between cloud and fog computing,'' \emph{International Journal of Distributed
  Sensor Networks}, vol.~13, no.~11, p. 1550147717742073, 2017.

\bibitem{tang2019offloading}
W.~Tang, X.~Zhao, W.~Rafique, L.~Qi, W.~Dou, and Q.~Ni, ``An offloading method
  using decentralized p2p-enabled mobile edge servers in edge computing,''
  \emph{Journal of Systems Architecture}, vol.~94, pp. 1--13, 2019.

\bibitem{dong2019joint}
C.~Dong and W.~Wen, ``Joint optimization for task offloading in edge computing:
  An evolutionary game approach,'' \emph{Sensors}, vol.~19, no.~3, p. 740,
  2019.

\bibitem{peng2021dosra}
Q.~Peng, C.~Wu, Y.~Xia, Y.~Ma, X.~Wang, and N.~Jiang, ``Dosra: A decentralized
  approach to online edge task scheduling and resource allocation,'' \emph{IEEE
  Internet of Things Journal}, vol.~9, no.~6, pp. 4677--4692, 2021.

\bibitem{polychronis2021safe}
G.~Polychronis and S.~Lalis, ``Safe optimistic path planning for autonomous
  drones under dynamic energy costs,'' in \emph{2021 IEEE International
  Intelligent Transportation Systems Conference (ITSC)}.\hskip 1em plus 0.5em
  minus 0.4em\relax IEEE, 2021, pp. 1927--1933.

\end{thebibliography}

\end{document}